\documentclass[10pt, a4paper]{article}
\usepackage[final]{lrec2026} 

\usepackage{booktabs}
\usepackage{color}
\usepackage{xurl}
\usepackage{hyperref}
\usepackage{multirow}
\usepackage{amssymb}
\usepackage{polyglossia}
\newfontfamily\nep{NotoSansDevanagari.ttf}[Script=Devanagari]

\title{NepTam: A Nepali-Tamang Parallel Corpus and Baseline Machine Translation Experiments}

\name{
    Rupak Raj Ghimire$^{1}$, 
    Bipesh Subedi$^{1}$,
    Balaram Prasain$^{2}$,
    Prakash Poudyal$^{1}$ \\
    {\bf \large Praveen Acharya$^{3}$,
    Nischal Karki$^{1}$,
    Rupak Tiwari$^{1}$,
    Rishikesh Kumar Sharma$^{1}$} \\
    {\bf \large Jenny Poudel$^{1}$,
    Bal Krishna Bal$^{1\ *}$\thanks{$^*$Corresponding author: bal@ku.edu.np}}
}
\address{
    $^{1}$ILPRL, Kathmandu University, Nepal\\
    $^{2}$Tribhuvan University, Nepal \\
    $^{3}$Dublin City University, Ireland\\
    \{rughimire, bipeshrajsubedi, prasainbalaram\}@gmail.com, prakash@ku.edu.np,\\ \{acharyaprvn, nischal3158, rupaktiwari18, rishi70612, jennypoudel100\}@gmail.com, bal@ku.edu.np
}

\abstract{
    Modern Translation Systems heavily rely on high-quality, large parallel datasets for state-of-the-art performance. However, such resources are largely unavailable for most of the South Asian languages. Among them, Nepali and Tamang fall into such category, with Tamang being among the least digitally resourced languages in the region. This work addresses the gap by developing \textit{NepTam20K}, a \textit{20K}  gold standard parallel corpus, and \textit{NepTam80K}, an \textit{80K} synthetic Nepali–Tamang parallel corpus, both sentence-aligned and designed to support machine translation. The datasets were created through a pipeline involving data scraping from Nepali news and online sources, pre-processing, semantic filtering, balancing for tense and polarity (in \textit{NepTam20K} dataset), expert translation into Tamang by native speakers of the language, and verification by an expert Tamang linguist. The dataset covers five domains: Agriculture, Health, Education and Technology, Culture, and General Communication. To evaluate the dataset, baseline machine translation experiments were carried out using various multilingual pre-trained models:mBART, M2M-100, NLLB-200, and a vanilla Transformer model. The fine-tuning on the NLLB-200 achieved the highest sacreBLEU scores of 40.92 (Nepali $\rightarrow$ Tamang) and 45.26 (Tamang $\rightarrow$ Nepali).
    \newline\newline \Keywords{Nepali-Tamang Parallel Corpus, Machine Translation, Nepali-Tamang Machine Translation, Low-resource Languages}
}

\begin{document}

\maketitleabstract

\section{Introduction}
The advancement in Artificial Intelligence (AI) and Machine Learning(ML), with the rise of Deep Learning techniques, has significantly transformed the domain of Natural Language Processing (NLP), including Machine Translation (MT). The improvement of results in contemporary MT systems is the driving force for researchers worldwide, pushing the boundaries of language technologies for resource-rich languages, thus opening the possibilities of extending the models trained in the resource-rich languages to low-resource languages. The Tamang language, native to the Sino-Tibetan language family and the Tibeto-Burman group, is spoken by over 1.42 million native speakers in Nepal~\cite{p7} and by the smaller communities in parts of Northeast India, including Sikkim, West Bengal and Assam. As of the 2021 census, Tamang is the fifth most spoken language of Nepal, making up about 4.88\% of the total population. Despite the large population of Tamang speakers, it remains a low-resource language in terms of the availability of digital footprint and datasets, along with computational tools. This has resulted  into challenges both in terms of linguistic preservation and technological inclusion of the language.
The lack of appropriate digital tools and technologies has led to poor usage of the language in digital communication, especially among the younger generations. Consequently, the language and its speakers have largely fallen behind in terms of utilizing the best benefits and potentials of modern internet technology, as the vast resources of knowledge are essentially in the English language globally, and the Nepali language in the case of Nepal. This dire state of the language can be addressed following sound approaches to data collection and adoption of modern ML techniques, including different MT deep learning architectures. In this context, we aim to develop a Tamang-Nepali parallel corpus and conduct baseline machine translation experiments to validate the effectiveness of our efforts in the development of the corpus.

Although Tamang traditionally uses the Tamyig (Tamang Alphabet) script \cite{ajitman1998}, very closely related to the Tibetan script, our corpus employs the Devanagari script, which is more widely adopted in Tamang literature. The former script, on the other hand, is largely confined to religious writings and thus lacks widespread use and practice. Besides Tamang, the Nepali language (which is written in the Devanagari script), an Indo-Aryan language of the Indo-European family, and the official language of Nepal, serves as a counterpart language in our parallel corpus. Nepali possess a relatively richer set of linguistic and digital resources compared to Tamang and is more widely spoken \cite{p7}. Hence, it has been appropriately selected as a suitable pivot language for developing an MT language pair in the project context. The development of a Tamang-Nepali parallel corpus thus strengthens linguistic resources for Tamang and contributes to the broader NLP ecosystem for underrepresented languages in Nepal.

The major contributions of this work include the development of a 20K gold-standard Nepali–Tamang parallel corpus aligned at the sentence level. Next, baseline machine translation experiments are conducted using state-of-the-art multilingual models, including mBART~\cite{p2}, M2M-100~\cite{p3}, NLLB-200~\cite{p4}, and the baseline Transformer~\cite{p1}, to establish performance benchmarks for future research on Nepali–Tamang translation. Finally, we also developed a \textit{NepTam80K} synthetic Tamang-Nepali parallel sentence-aligned corpus built by the best-performing model from our experiments. 

The rest of the paper is organized as follows: Section 2 presents related works, followed by the corpus development methodology in Section 3. Sections 4 and 5 present the details of the experiments and a discussion of the results. Finally, the paper concludes with conclusion and future work in Section 6.

\section{Related works}
NLP for low-resource and endangered languages is getting increasing focus as researchers around the world are working to bridge the digital gap in linguistically diverse regions.  But it should be noted that many South Asia languages remain under-represented due to limited resources. ~\citet{p8} notes the lack of standard datasets, benchmarks, and corpora, especially for Tibeto-Burman languages, which pose tonal, morphological, and script challenges. There have been some efforts like IndoLib~\cite{p9} trying to address this, though progress depends on data quality and availability. Amidst these hurdles, research is expanding in languages such as Nepali, Tamang, Sinhala, Dzongkha, Maithili, Assamese, and Kashmiri. A substantial number of works are underway in the Nepali language for application like Machine Translation~\cite{p10}, Speech Recognition~\cite{nepconformer, p11, ghimire2023active, ghimire2025improving}, Sentiment Analysis~\cite{p14}, NER and POS Tagging~\cite{p12}, and Image/Video Captioning~\cite{p13,p41}. Nevertheless, development remains constrained by limited linguistic resources, making gold-standard corpora and parallel datasets crucial. In this vein, ~\citet{p36} highlights persistent issues in corpus creation, including data scarcity, domain bias, code-mixing, and linguistic diversity, though growing initiatives continue to improve resource availability.

Several initiatives have expanded resources for Indic and other low-resource languages, enhancing their linguistic and technological value. ~\citet{p16} introduced Samanantar, a 49.7M-sentence parallel corpus for 11 Indic languages, built from OPUS, localization materials, religious texts, subtitles, and web-mined pairs, filtered using semantic similarity and human evaluation, though issues like noise and uneven coverage remained. ~\citet{p15} developed the IndicNLP Corpus, containing 2.7B words from 10 Indic languages via web crawls of news and Wikipedia. ~\citet{p27} advanced this with IndicTrans2, creating the Bharat Parallel Corpus Collection with 230.5M bitext pairs across 22 Indic languages (including Nepali), combining 2.2M human-translated gold pairs with web data and quality filtering. Similarly, ~\citet{p18} built the COMPARA English–Portuguese corpus through OCR, text cleaning, and formatting for structured retrieval.

Recently,~\citet{p19} combined translation (50\%), crawling (42\%) and sourcing of existing parallel data (8\%)  to construct an English–Tshivenda corpus. The corpus was also cleaned and filtered to remove overlapping or low-quality sentences. These cleaned sentences were translated into Tshivenda by linguists and a language expert. In ~\citet{p20}, the data collection process for the Uzbek–Kazakh parallel corpus was carried out in three stages. In the first stage, a small set of existing parallel resources was collected, totalling 138 sentence pairs. The second stage focused on automatic alignment, where parallel texts from bilingual web news articles and translated literature were crawled, cleaned, and aligned. Finally, the third stage involved large-scale manual translation of 100,000 Uzbek sentences into Kazakh by bilingual students, followed by expert cross-checking to ensure translation accuracy.

In the Nepali context, several monolingual Nepali corpora have been created such as IRIIS-RESEARCH/Nepali\-Text\_Corpus~\cite{p38}, OSCAR dataset ~\cite{p39}, and NepBERTa~\cite{p40}. These datasets contain plain text without sentence-level categorization, while our objective requires each sentence to be assigned a specific category.
Moreover, data collection methodologies reveal persistent gaps in scale, quality, and diversity.  For instance, ~\citet{p43} compared phrase-based SMT and RNN-based NMT for English-Nepali using a small 6.5K parallel corpus borrowed from the Nepali National Corpus (NNC)~\cite{yadava2008construction}, which was augmented with linguist-collected sentences. Similarly, ~\citet{p44} augmented an English-Nepali corpus to 1.8M sentences by cleaning available data corpus such as GNOME/Ubuntu/KDE~\cite{tiedemann2012parallel}, in which they reduced the data from 500K to 58K by manual editing. They also created a synthetic corpus of about 1.6 million parallel sentences using back translation.

Domain-specific efforts, such as~\citet{p10}, built a 125K English-Nepali legal corpus. The paper highlights some gaps, including the absence of linguistic verification, lack of diverse data, domain bias, etc. 

Most relevant to our study, ~\citet{p6} constructed a 15,000 sentence level bilingual corpus for Nepali-Tamang. The data collection process involved extracting raw texts from diverse sources such as child storybooks, Tamang language magazines, and spoken dialogues, followed by collaboration with Tamang linguists for accurate translation and sentence alignment.

The development of translation systems is increasing day by day. In the South Asian context, it has been proven through the project such as Ai4Bharat ~\cite{p27}, Bhashini ~\cite{p28}, and Aksharantar ~\cite{p29} for Indic languages. ~\citet{p43} has worked with an English-Nepali parallel corpora of size 6.5K, comparing the results of SMT and NMT-based systems in the case of low resource availability. It is found that SMT based systems tend to outperform NMT-based systems on the BLEU evaluation metric, with a respective BLEU score of 5.27 and 3.28 in the direction of English to Nepali.

Similarly, work done by ~\citet{p10} presents a Bidirectional English-Nepali Machine Translation System for the Legal Domain where they utilized a Neural Machine Translation (NMT) System with an encoder-decoder architecture, designed for legal Nepali-English translation. Leveraging a custom-built legal corpus of 125,000 parallel sentences, their system achieved the BLEU scores of 7.98 in (Nepali $\rightarrow$ English) and 6.63 (English $\rightarrow$ Nepali) direction. Another work by~\citet{kritiNemkul2021}  includes the development of  English to Nepali sentence translation using long short-term memory (LSTM) cells, in its encoder and decoder with attention. The LSTM cells with two layers of neural network and 256 hidden units were found to have the highest BLEU score of 8.9.
~\citet{p6} developed a bidirectional Transformer-based NMT system for a Nepali-Tamang MT translation, achieving BLEU scores of 27.74 (Nepali $\rightarrow$ Tamang) and 23.74 (Tamang $\rightarrow$ Nepali), contributing to the inception of formal work in Nepali and Tamang translation.

Building on the broader landscape of multilingual NMT, several studies have implemented pretrained multilingual models for low-resource Indian and related languages, including NLLB-200, mBART, and M2M-100 ~\cite{p16, p30, p31, p32, p33, p35, p27}. However, these models have not yet been explored in the context of the Nepali–Tamang language pair. ~\citet{p6} primarily applied Transformer-based models trained on small and domain-neutral datasets without expert review or linguistic diversity and inclusiveness. To address this gap, we compile a 20K Nepali–Tamang parallel corpus featuring linguistic diversity, expert validation, and broad domain coverage, and evaluate state-of-the-art multilingual pretrained models to assess their performance on this underrepresented pair.

\section{Corpus Development and Data Preparation}\label{section3}
\begin{figure*}[t]
  \centering
  \includegraphics[scale=0.9,keepaspectratio,trim=4mm 4mm 4mm 4mm, clip]{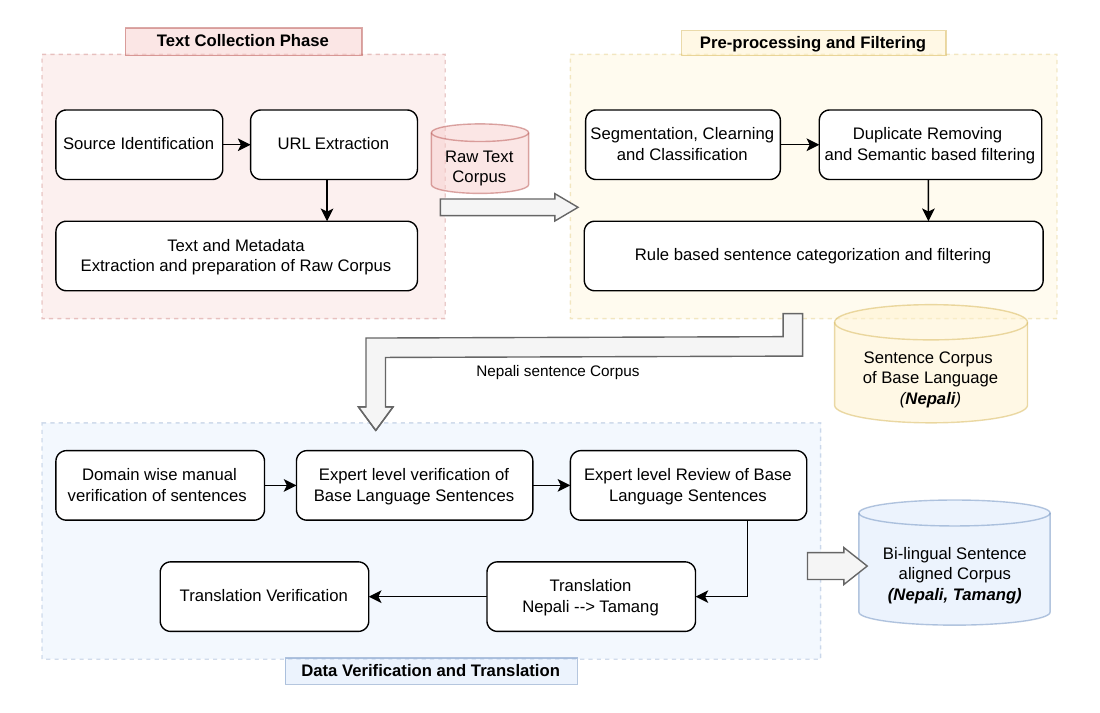}
  \caption{Overall procedure of the proposed Nepali–Tamang Machine Translation resource development}
  \label{fig:overall_procedure}
\end{figure*}

\subsection{Text and Metadata Extraction}
In recent years, the digital transformation has led to a significant increase in Nepali-language content across various websites. Our goal lies in using this content to prepare the translation dataset in five categories: \textit{(1) Agriculture} \textit{(2) Health} \textit{(3) Education and Technology} \textit{(4) Culture, Tourism and Society}, and \textit{(5) General Communication} (derived from Stories, Literature, Arts). Some related categories were combined to create broader groups that capture diverse aspects of data and address the issue of data scarcity within individual domains. In order to meet the requirements, we created a text corpus maintaining the metadata, including category, published date, author, keywords, etc. During this phase, we collected data from the popular news portals of Nepal and stocked approximately 10 GB of text with their associated metadata.

The resource development process of our 20K parallel corpus starts from text collection, followed by pre-processing and finally data verification and translation. The overall process of resource development is summarized in Figure \ref{fig:overall_procedure}.

Our initial approach involved collecting article data directly through Common Crawl’s index\footnote{\url{https://commoncrawl.org/get-started}}. We collected 4.35 million de-duplicated URLS. The contents of these URLs are then extracted using a general parsing script that works across all 154 top-level domains. While it is generally a straightforward process to detect the article body and title, designing a general script to extract metadata, particularly the article category, is much more challenging. Each domain uses unique tags, attributes, or structures to represent such information within its HTML document. To address this, a layered funnel architecture based on conditional logic is implemented to handle domain-specific variations effectively. This approach successfully captures the article categories for the majority of the source domains.  Moreover, the extracted categories extend beyond the intended scope, as the study focuses on five primary categories.

\subsection{Preprocessing and Filtering}

\subsubsection{Segmentation, cleaning, and classification}

In the first step of the data pre-processing pipeline, the articles are converted into sentences using the tokenizer from the Indic NLP library. The sentence cleaning pipeline is made to refine and standardize Nepali text data for further processing.
Initially, the sentences containing English words are removed with the help of language detection tools to ensure the resulting sentences are explicitly in Nepali. Then, the regular expressions (regex) are applied to eliminate unwanted information such as numbers (0–9,{\nep ०-९}), textual advertisements({\nep "सूचना तथा सुझाव"} (Notice and Suggestions)), excessive special characters, and other unwanted phrases. Irrelevant symbols like ellipses (...), outer quotation marks, and unwanted leading characters such as ``-'', ``:'', and ``\_'' are also removed. Common typographical errors are spotted and corrected using regex-based replacements (e.g., {\nep "न्"} instead of mistyped {\nep "न््"}) or manual intervention. In order to maintain uniform formatting, unnecessary spaces and commas before the period {\nep "।"} are removed. Additionally, leading location markers (e.g., {\nep "काठमाडौं:"} (``Kathmandu:"),{\nep "पोखरा –"} (``Pokhara -'') are discarded to keep only the meaningful sentence content. The sentences that do not end with verbs are discarded, as they often represent incomplete or irrelevant fragments. Additionally, we used the regular expression patterns to identify verbs derived from~\citet{p45}. We also matched non-verbal forms that share similar endings, potentially leading to the inclusion of a few sentences that do not actually end with verbs. The sentences are then checked for duplication. Some examples are listed in Table~\ref{tab:cleaning}.

\begin{table}[h!]
\centering
\begin{tabular}{cl}
\hline
\multirow{3}{*}{Before} & {\nep काठमाडौं: आजको मौसम राम्रो छ} (Nepali) \\
                        & kathmandu: aajako mausam ramro chha   \\
                        & Kathmandu: Today's weather is good    \\ \cline{2-2}
\multirow{3}{*}{After}  & {\nep आजको मौसम राम्रो छ।} (Nepali)                  \\
                        & aajako mausam ramro chha.              \\
                        & Today's weather is good.               \\ \hline
\multirow{3}{*}{Before} & {\nep \_घरमा खाना पकाउँदैछ , ।} (Nepali)    \\
                        & \_gharma khana pakaudaichha , |        \\
                        & \_Making food at home , .              \\ \cline{2-2}
\multirow{3}{*}{After}  & {\nep घरमा खाना पकाउँदैछ।} (Nepali)                 \\
                        & gharma khana pakaudaichha|             \\
                        & Making food at home.                   \\ \hline
\end{tabular}%
\caption{Example of Text Before and After pre-processing}
\label{tab:cleaning}
\end{table}

To ensure the systematic organization of category-specific articles, all collected sentences are classified into five primary categories based on their semantic relevance and associated metadata. Due to inconsistencies in category naming across different sources, a regex based keyword mapping technique is used to regularize these categories.
Moreover, to include sentences of different lengths, we classified them into the following types:
\begin{itemize}
    \item Short Sentences: 3-7 words
    \item Medium Sentences: 8-15 words
    \item Long Sentences: 16-21 words
    \item Very Long Sentences: > 21 and <40 words
\end{itemize}
The sentence lengths are arbitrarily selected as per the suggestion of the Nepali linguist expert. Very long sentences are included in a limited number to maintain diversity in terms of sentence length. As far as the data borrowed from ~\citet{p6}, we did some basic cleaning, de-duplication, and handled punctuation issues to some extent. A small portion of this corpus (around 350 sentences) contains only two words. These very short sentences are retained to preserve coverage of the source material.

\subsubsection{Semantic filtering}
The data collected from various news domains may contain semantic duplicates resulting from the repetition of identical content across multiple sources.
To address this issue, the semantic similarity between each pair of sentences is computed, and in cases where the similarity score exceeds the threshold of 0.8 (as determined by a Nepali linguist through sampling), only one of the similar sentences is retained. This process involves $n \times n$ pairwise comparisons within the category, where $n$ denotes the number of sentences. Text embedding models, jangedoo/all-MiniLM-L6-v2-nepali\footnote{https://huggingface.co/jangedoo/all-MiniLM-L6-v2-nepali} and LaBSE~\cite{p42}, which are fine-tuned on Nepali text, are used to calculate the similarity vectors followed by FAISS (Facebook AI Similarity Search)\footnote{https://faiss.ai/} for semantic filtering. It helps to search for similar vectors efficiently in large datasets. This approach ensures the inclusion of sentences with diverse levels of semantic similarity within each category. However, semantic filtering was not applied to the parallel corpus from~\citet{p6} to avoid removing valuable translation pairs and compromising its representativeness.

\subsubsection{Rule-based classification of tense \& polarity}
Each sentence is classified based on tense and polarity using a rule-based classifier that analyzes the morphological inflection patterns of verbs in the Devanagari script. Following the approach of ~\citet{p45}, tense is categorized as Past or Non-Past, and polarity as Affirmative or Negative. Table \ref{tab:tenseclassifier} and Table \ref{tab:polarityclassifier} show the list of patterns(verbs) for tense and polarity classification used in this study. Each sentence is first cleaned and tokenized to identify the final verb or verb phrase, which contains the tense and polarity information. Then the system checks the endings of verbs (inflections) against a set of regular expression patterns that show how common Nepali verbs are conjugated. The presence of auxiliary verbs like {\nep "थियो/थिए"} (past) and {\nep "छ/छन्"} (non-past), as well as past-tense markers like {\nep "एँ", "यो", "एको"}, to name a few, determines the tense. Polarity detection also depends on being able to spot negative prefixes like {\nep "न"} and negative inflections such as {\nep "छैन," "हुँदैन," "थिएन"}. When auxiliary verbs are missing, the classifier uses suffix-based inflection rules to figure out the tense, giving longer pattern matches more weight for accuracy. Although we aimed to exclude sentences without a verb at the end, some non-verb words may exhibit similar patterns or inflections, causing such sentences to remain in the dataset.

\begin{table}[h!]
    \centering
    \begin{tabular}{p{2cm}p{5cm}}
        \toprule
            \textbf{Tense} & \textbf{Pattern (Verb)} \\
        \midrule
        Non Past & 
            {\nep छु, छौँ, छस्, छेस्, छौ, छ, छे, छन्‌, छिन्‌, छौं, दिनँ, दैनौँ, दैनस्, दैनौ, दिनौ}\\
            \hline
        Past & 
            {\nep एँ, यौँ, यौं, इस्, यौ, यो, ई, ए, इन्, इनँ, एनौ, इनस्, इनौ, एन, इन, एनन्, इनन्} \\
            \hline
        Unknown & Does not end with the pattern\\
        \bottomrule
    \end{tabular}
     \caption{Tense Classification based on Verb Patterns ~\cite{p45}}
     \label{tab:tenseclassifier}
\end{table}

\begin{table}[h!]
    \centering
    \begin{tabular}{p{2cm}p{5cm}}
        \toprule
            \textbf{Polarity} & \textbf{Pattern (Verb)} \\
        \midrule
        Affirmative &
            {\nep छु, छौँ, छस्, छेस्, छौ, छ, छे, छन्‌, छिन्‌, एँ, यौँ, यौं, इस्, यौ, यो, ई, ए} \\
            \hline
        Negative &
            {\nep दिनँ, दैनौँ, दैनस्, दैनौ, दिनौ, इनौ, एन, इन, एनन्, इनन्}\\
            \hline
        Unknown & Does not end with the pattern \\
        \bottomrule
    \end{tabular}
    \caption{Polarity Classification based on Verb Patterns~\cite{p45}}
    \label{tab:polarityclassifier}
\end{table}

To achieve balanced distributions among different linguistic variables such as sentence type, similarity category, tense, and polarity and to ensure linguistic diversity within the dataset, the data is distributed according to established patterns. 
Short, medium, and long sentences are represented in proportions that favour medium-length sentences while maintaining sufficient examples of short and long forms. Similarly, sentences are organized to include a mix of low, medium, and high similarity. The tense and polarity are also intended to be evenly distributed.
Given that the dataset may not always perfectly satisfy these targets, proportions are dynamically adjusted to maximize diversity. Moreover, existing data from~\citet{p6} being limited and specific, does not fully follow the target distributions, but it adds valuable translation examples and improves coverage.

\subsection{Translation and Verification}
After pre-processing and filtering, the data are manually reviewed and verified by a linguist expert. Sentences that are inconsistent in writing format or semantic meaning were excluded. After that these sentences are assigned to the experts proficient in the Tamang language capable of accurately conveying contextual and semantic nuances. 

The translation team consisted of five trained translators with formal training in linguistics and a strong command of both written and spoken Tamang and Nepali. Tamang translations were carried out by linguistics graduates, while the Nepali source text and corresponding English translations were finalized by a Professor and Senior Computational Linguist. The Tamang translations were further verified and finalized by a senior expert holding a Master’s degree in Linguistics, a long-standing social worker and activist in the Tamang language community, author of more than 70 books on the Tamang language, grammar, and culture, and former Editor-in-Chief of a Tamang-language magazine for over 20 years.

Both Nepali and Tamang languages are written in the Devanagari script for consistent orthographic representation. Following the translation process, a group of linguistic experts verified the translated corpus to ensure consistency, accuracy, and quality. Furthermore, the translators and reviewers were supported by members of the Tamang community, ensuring linguistic reliability and cultural authenticity throughout the process. 

The parallel corpus developed by~\citet{p6} was reviewed by a Nepali linguistic expert from our team, as their original work involved only Tamang linguists. We selected 10K sentences out of 15K sentences after the manual review process mentioned above.

\subsection{Dataset Details}
\begin{table*}[h!] 
\centering 
\begin{tabularx}{0.8\textwidth}{p{4cm}XXX} 
    \toprule 
    \textbf{Category} & \multicolumn{2}{c}{\textbf{NepTam20K}} & \textbf{NepTam80K} \\ 
    \cmidrule(lr){2-3}
    & \textbf{Train} & \textbf{Test} & \\ 
    \midrule 
    \textit{Sentence Length} & & & \\
    \midrule
    \hspace{0.5cm}Short & 5,863 (39.1\%) & 1,954 (39.1\%) & 15,941 (19.9\%)\\
    \hspace{0.5cm}Medium & 6,793 (45.3\%) & 2,266 (45.3\%) & 52,122 (65.2\%)\\ 
    \hspace{0.5cm}Long & 1,552 (10.3\%) & 518 (10.4\%) & 10,640 (13.3\%)\\ 
    \hspace{0.5cm}Very Long & 823 (5.5\%) & 262 (5.2\%) & 1,396 (1.7\%)\\ 
    \midrule
    \textit{Tense} & & & \\
    \midrule
    \hspace{0.5cm}Non-Past & 6,781 (45.2\%) & 2,282 (45.6\%) & 55,542 (69.4\%)\\
    \hspace{0.5cm}Past & 6,436 (42.9\%) & 2,131 (42.6\%) & 24,295 (30.3\%)\\
    \midrule
    \textit{Polarity} & & & \\
    \midrule
    \hspace{0.5cm}Affirmative & 11,939 (79.6\%) & 3,978 (79.6\%) & 74,871 (93.6\%)\\ 
    \hspace{0.5cm}Negative & 1,278 (8.5\%) & 435 (8.7\%) & 4,966 (6.2\%)\\ 
    \hspace{0.5cm}Unknown & 1,784 (11.9\%) & 587 (11.7\%) & 262 (0.3\%)\\ 
    \midrule 
    \textbf{Total Sentences} & 15,000 & 5,000 & 80,099\\ 
    \bottomrule
\end{tabularx} 
\caption{Category-wise Data Distribution in \textit{NepTam20K} and \textit{NepTam80K} Dataset.} 
\label{tab:dataset-details}
\end{table*}

Throughout this work, we developed two primary datasets for Nepali–Tamang translation: a gold-standard parallel dataset \textit{NepTam20K} and a synthetic dataset \textit{NepTam80K}. The gold-standard dataset consists of 20K sentence-aligned and translated parallel sentences. Half of this data (10K sentences) is created following our methodology, ensuring coverage across five categories, with 2K sentences in each category. The remaining 10K sentences were sourced from the parallel corpus of~\citet{p6} and underwent cleaning, de-duplication, and manual review by a Nepali linguist to ensure consistency and quality.

Following the strategies for corpus development for low-resource languages described in ~\citet{bal2024strategies}, we augment the dataset to create a synthetic parallel corpus, \textit{NepTam80K}, consisting of 80K sentence pairs through translation of Nepali sentences into Tamang using our best-performing translation model. Its tense/polarity distribution reflects the underlying monolingual data and model generation. As shown in Table \ref{tab:dataset-details}, NepTam80K is skewed toward non-past and affirmative forms, mainly because the scraped monolingual data were not balanced and contained a higher proportion of such forms compared to other tense and polarity categories. 

Table ~\ref{tab:dataset-details} shows the details of the final corpus.

\section{Experiments}
\subsection{Language Tagging Strategy}
Tamang is not included in the pre-trained NLLB, M2M, or mBART vocabularies, so we used the Hindi language tag to represent Tamang text during fine-tuning. We emphasize that all training data remained in Tamang, and the model was fine-tuned on genuine Tamang sentences. Using the Hindi tag was a practical workaround to enable training with pre-trained multilingual models that lack a dedicated Tamang token. While this does not perfectly capture language-specific features, it allowed effective fine-tuning and transfer of multilingual knowledge to the Tamang–Nepali translation task.
Existing research highlights the utility of leveraging high-resource language tags as a technical workaround for low-resource ones ~\cite{p47, p48}.
\citet{p47} shows that transliterating low-resource text into the script of a Related Prominent Language (RPL), such as Hindi, enhances performance. Likewise, \citet{p48} claims that Indic languages with orthographic and phonetic similarity benefit from shared embeddings, enabling transfer to unseen languages.

\subsection{Training and Evaluation}

For the experiment, three multilingual MT models - mBART, M2M-100, and NLLB-200 were fine-tuned using the Hugging Face library. On the other hand, a baseline standard Transformer was trained from scratch using the fair-seq library~\cite{ott2019fairseq}. The models were trained on 15K and tested on 5K sentence pairs from the NepTam20K dataset. The training and fine-tuning hyperparameters used are listed in Table~\ref{tab:model_config}. Standard text pre-processing and tokenization pipelines were employed based on the respective original model's tokenizer.

\begin{table}[t]
\centering
\footnotesize
\begin{tabular}{llc}
\hline
\textbf{Model} & \textbf{Hyperparameter} & \textbf{Value} \\
\hline
$NepTam_{mBART}$  & Epochs             & 7 \\
          & Batch size         & 8 \\
          & Grad. accumulation & 2 \\
          & Learning rate      & 7e-5 \\
          & Weight decay       & 0.01 \\
\hline
$NepTam_{M2M}$   & Epochs             & 5 \\
          & Batch size         & 8 \\
          & Grad. accumulation & 2 \\
          & Learning rate      & 7e-5 \\
          & Weight decay       & 0.01 \\
\hline
$NepTam_{NLLB}$  & Epochs             & 5 \\
          & Batch size         & 16 \\
          & Learning rate      & 5e-4 \\
          & Dropout            & 0.3 \\
\hline
$NepTam_{Transformer}$ & Encoder layers  & 5 \\
            & Decoder layers  & 5 \\
            & Embedding size  & 512 \\
            & FFN dimension   & 2048 \\
            & Attention heads & 8 \\
            & Epochs          & 50 \\
            & Learning rate   & 5e-4 \\
            & Dropout         & 0.3 \\
            & Weight decay    & 1e-4 \\
\hline
\end{tabular}
\caption{Training and fine-tuning configurations for all models on NepTam20K.}
\label{tab:model_config}
\end{table}

For convenience, we use the following naming conventions to denote specific models:
\begin{itemize}
    \item $NepTam_{M2M}$ : \small fine tuned M2M-100 model
    \item $NepTam_{mBART}$ : \small fine tuned mBART-50 model
    \item $NepTam_{NLLB}$ : \small fine tuned NLLB-200 model
    \item $NepTam_{Transformer}$ : \small vanilla transformer model
\end{itemize}

To further evaluate the model's ability to leverage a larger dataset, re-training was performed on a NepTam80K pair where Tamang sentences were synthetically generated by translating Nepali sentences using the previously trained best-performing checkpoint ($NepTam_{NLLB}$). Using this synthetic parallel data, the models were re-trained from their previous best checkpoints. This procedure allowed for assessing the impact of synthetic data augmentation on translation quality and the model's capacity to generalize beyond the limited original dataset.

All the models were evaluated using sacreBLEU, chrF, chrF++, METEOR, and COMET. Use of these complementary metrics together provides a well-balanced evaluation across both translation directions.

\begin{table*}[!ht]
    \centering
    \begin{tabularx}{\textwidth}{p{3.0cm} p{1.7cm} p{1.5cm} c c c c c}
        \toprule
            \textbf{Model} & \textbf{Parameter Size} & \textbf{Direction} &
            \textbf{sacreBLEU} & \textbf{chrF} & \textbf{chrF++} &
            \textbf{METEOR} & \textbf{COMET} \\
        \midrule
            $NepTam_{M2M}$ & 418M  & 
                ne$\rightarrow$tam & 
                    40.24 & 73.30 & 69.27 & 60.81 & 75.67 \\
                & & tam$\rightarrow$ne & 
                    42.73 & 72.21 & 68.74 & 61.07 & 79.13 \\
        \midrule
            $NepTam_{NLLB}$  & 600M  & 
                ne$\rightarrow$tam &
                    \textbf{40.92} & \textbf{73.98} & \textbf{69.94} & \textbf{61.44} & \textbf{75.89} \\
                & & tam$\rightarrow$ne & 
                    \textbf{45.26} & \textbf{73.71} & \textbf{70.30} & \textbf{62.31} & \textbf{80.19} \\
        \midrule
            $NepTam_{mBART}$  & 610.9M  & 
                ne$\rightarrow$tam & 
                    40.14 & 72.91 & 68.92 & 60.46 & 75.60 \\
                & & tam$\rightarrow$ne & 
                    42.96 & 71.65 & 68.25 & 60.43 & 79.04 \\
        \midrule
            $NepTam_{Transformer}$ & 49.07M & 
                ne$\rightarrow$tam & 
                    37.71 & 71.71 & 67.74 & 58.20 & 75.00 \\
                & & tam$\rightarrow$ne & 
                    38.01 & 69.89 & 66.30 & 57.60 & 77.37 \\
        \bottomrule
    \end{tabularx}
    \caption{Performance of translation models trained on NepTam20K (Train), evaluated on NepTam20K (Test) set for Nepali–Tamang (ne$\rightarrow$tam) and Tamang–Nepali (tam$\rightarrow$ne) directions}
    \label{tab:translation-results}
\end{table*}

\begin{table*}[ht!]
    \centering
    \begin{tabularx}{\textwidth}{p{5.0cm} c c c c c c c}
        \toprule
            \textbf{Model} & \textbf{Direction} &
            \textbf{sacreBLEU} & \textbf{chrF} & \textbf{chrF++} &
            \textbf{METEOR} & \textbf{COMET} \\
        \midrule
            $NepTam_{M2M}$ & 
                ne$\rightarrow$tam & 
                    40.56 & 73.70 & 69.62 & 60.55 & 75.68 \\
                & tam$\rightarrow$ne & 
                    44.03 & 72.83 & 69.34 & 60.50 & 79.28 \\
        \midrule
            $NepTam_{NLLB}$ & 
                ne$\rightarrow$tam &
                    41.79 & 74.58 & 70.60 & 62.19 & 76.12 \\
                & tam$\rightarrow$ne & 
                    48.28 & 75.57 & 72.31 & 64.25 & 81.26 \\
        \midrule
            $NepTam_{mBART}$ & 
                ne$\rightarrow$tam & 
                    41.44 & 74.07 & 70.10 & 61.21 & 75.83 \\
                &  tam$\rightarrow$ne & 
                    46.68 & 74.59 & 71.17 & 61.96 & 80.70 \\
        \midrule
            $NepTam_{Transformer}$ & 
                ne$\rightarrow$tam & 
                    37.72 & 72.13 & 68.01 & 57.13 & 74.61 \\
                & tam$\rightarrow$ne & 
                    38.24 & 69.32 & 65.60 & 54.26 & 75.76 \\
        \bottomrule
        
    \end{tabularx}
    \caption{Performance of translation models trained on NepTam80K(Train), evaluated on NepTam20K Test set for Nepali–Tamang (ne$\rightarrow$tam) and Tamang–Nepali (tam$\rightarrow$ne) directions}
    \label{tab:translation-results-80k}
\end{table*}

\section{Result and Discussion}

We performed experiments on various pre-trained models and summarized the results in Table \ref{tab:translation-results}. Among the evaluated models, $NepTam_{NLLB}$ outperformed all other models across sacreBLEU, chrF, chrF++, METEOR, and COMET metrics in both translation directions for the Tamang–Nepali language pair. For the Nepali $\rightarrow$ Tamang direction, it reached a sacreBLEU score of 40.92 and, while in the Tamang $\rightarrow$ Nepali direction, it reached a sacreBLEU score of 45.26. Our models achieve competitive scores on
several metrics within this new test set. This demonstrates the model's strong multilingual capability and adaptability to the low-resource Nepali–Tamang pair. On the other hand, the Transformer baseline produced the lowest scores across all metrics, confirming the importance of leveraging large-scale multilingual pre-training for under-represented languages. Models such as $NepTam_{M2M}$ and $NepTam_{mBART}$ also performed competitively, indicating that multilingual pre-training helps in bridging the linguistic gap, even when direct training data for the target pair is scarce.

Across all evaluated models, translation in the Tamang $\rightarrow$ Nepali direction demonstrated superior performance compared to Nepali $\rightarrow$ Tamang. The results further highlight that fine-tuning pre-trained models, even with limited data, can yield substantial improvements in translation quality. These findings provide a solid baseline for future research on Nepali-Tamang translation, where model adaptation and data augmentation techniques could be explored to further enhance performance.

Furthermore, we also performed fine-tuning of the $NepTam_{M2M}$, $NepTam_{NLLB}$, $NepTam_{mBART}$ and $NepTam_{Transformer}$ models using the synthetic \textit{NepTam80K} dataset. We then evaluated these models on the \textit{NepTam20K} (Test) dataset. We observed that the performance of all models improved except for the $NepTam_{Transformer}$, which exhibited similar results as in baseline experiments, as shown in Table \ref{tab:translation-results-80k}. $NepTam_{NLLB}$ performed better than other models with BLEU scores of 41.79 in (Nepali $\rightarrow$ Tamang) and 48.28 in (Tamang $\rightarrow$ Nepali) direction.

\section{Conclusion}

This work introduces the first large-scale Nepali–Tamang parallel dataset: \textit{NepTam20K}, a gold-standard corpus and \textit{NepTam80K}, an 80K synthetic corpus, offering extensive domain diversity and comprehensive coverage of linguistic phenomena, thereby providing a richer and more representative resource for machine translation research. Both datasets are sentence-aligned and were used to fine-tune state-of-the-art multilingual MT models such as \textit{mBART}, \textit{M2M-100}, and \textit{NLLB-200}, on which we conducted baseline experiments to demonstrate their effectiveness. We achieved promising results across standard translation metrics. The corpus creation pipeline from multilingual data scraping, pre-processing, semantic filtering, and linguistic annotation to manual translation ensures data quality, linguistic diversity, and domain balance. The experimental findings confirm that multilingual pretrained models can effectively adapt to low-resource settings, offering a strong baseline for Tamang-Nepali machine translation and a foundation for future NLP applications involving underrepresented languages.

Future work can focus on expanding the corpus via community-driven translation efforts, back-translation and data augmentation to improve the model generalization. Further gains may be achieved through domain-specific fine-tuning and exploring cross-lingual transfer learning from related Tibeto-Burman or Indo-Aryan languages. It is also worth exploring the use of the instruction-tuned LLMs for Nepali–Tamang translation and for controlled synthetic data generation. The released data splits also enable future comparison of the dataset and translation performance with multilingual MT baselines. Moreover, statistical significance testing should be conducted in future to confirm whether the observed gains, especially the smaller ones, are reliable.

\paragraph{Dataset and Models:} 
The NepTam Corpus, along with the associated model training notebooks, is publicly available for research purposes. The complete dataset and experimental notebooks can be accessed at the following repository: \url{https://github.com/ilprl/NepTam-A-Nepali-Tamang-Parallel-Corpus-and-Baseline-Machine-Translation-Experiments}.

\section{Acknowledgments}
This research was funded by Google through the 2024 Google Academic Research Award (GARA) under the Society-Centered AI initiative and Taighde Éireann – Research Ireland under Grant No. 18/CRT/6223.

\section{Limitations}
The proposed approach is primarily based on fine-tuning existing multilingual machine translation models, and thus the main contribution is the construction of the Nepali–Tamang parallel corpus rather than methodological innovation. In addition, the system uses Hindi language tag for Tamang and incorporates synthetic data, automatic metrics such as sacreBLEU and chrF++ may not fully capture true translation quality. Human evaluation and qualitative error analysis were not included and will be considered in future work.

\section{Ethical considerations}
This study develops a Nepali–Tamang machine translation system using a corpus constructed from publicly accessible online sources. The Nepali data were carefully curated by a senior linguist to ensure quality and appropriateness. Tamang translations were produced by a dedicated translator/linguist team. Only publicly available text was collected, and no private or personally identifiable information was intentionally included. Nonetheless, web-sourced data may reflect domain imbalance or embedded societal biases. We acknowledge potential risks of mistranslation, semantic distortion, or cultural misrepresentation, particularly in low-resource settings. Accordingly, the system should not be deployed in high-stakes contexts without human oversight. The primary objective of this work is to promote linguistic inclusion and digital support for under-resourced language communities.

\section{Bibliographical References}\label{sec:reference}

\bibliographystyle{lrec2026-natbib}
\bibliography{lrec2026}

\end{document}